# Developing large language model for BIM-based design with domain-specific benchmark and dataset

Jia-Rui Lin [a,b], Yun-Hong Cai [a,b], Xiang-Rui Ni [a,b,c], Shaojie Zhou [a,b], Peng Pan [a,b]

[a] Department of Civil Engineering, Tsinghua University, Beijing, China, 100084

[b] Key Laboratory of Digital Construction and Digital Twin, Ministry of Housing and Urban-Rural Development, Beijing, China, 100084

[c] PetroChina Planning & Engineering Institute, Beijing, China, 100083

**Abstract**

As the construction industry advances toward digital transformation, BIM (Building Information Modeling)-based design has become a key driver supporting intelligent construction. Despite Large Language Models (LLMs) have shown potential in promoting BIM-based design, the lack of specific datasets and LLM evaluation benchmarks has significantly hindered the performance of LLMs. Therefore, this paper addresses this gap by proposing: 1) an evaluation benchmark for BIM-based design together with corresponding quantitative indicators to evaluate the performance of LLMs, 2) a method for generating textual data from BIM and constructing corresponding BIM-derived datasets for LLM evaluation and fine-tuning, and 3) a fine-tuning strategy to adapt LLMs for BIM-based design. Results demonstrate that the proposed domain-specific benchmark effectively and comprehensively assesses LLM capabilities, highlighting that general LLMs are still incompetent for domain-specific tasks. Meanwhile, with the proposed benchmark and datasets, Qwen-BIM is developed and achieves a 21.0% average increase in G-Eval score compared to the base LLM model. Notably, with only 14B parameters, performance of Qwen-BIM is comparable to that of general LLMs with 671B parameters for BIM-based design tasks. Overall, this study develops the first domain-specific LLM for BIM-based design by introducing a comprehensive benchmark and high-quality dataset, which provide a solid foundation for developing BIM-related LLMs in various fields.

# 1 Introduction

Building Information Modelling (BIM) technology has been proposed and rapidly applied in the field of construction since 2000. As a unified digital model for engineering projects, BIM enables cross-disciplinary collaboration, supports coordinated management across project stages, and integrates comprehensive lifecycle information to facilitate conflict detection and data-driven decision-making, thereby significantly reducing errors and improving overall project quality and efficiency(Eastman et al., 2009; Nguyen and Adhikari, 2023). BIM-based design not only streamlines complex engineering tasks, such as automated clash resolution(Akponeware and Adamu, 2017), intelligent site layout planning(Kumar and Cheng, 2015), and structural performance optimization(Mangal et al., 2021), but also serves as the foundation for sustainable design, facilitating precise energy analysis and lifecycle performance assessment to meet increasingly stringent green building standards(Jalaei F, 2014). As the industry shifts toward digital transformation, BIM-based design has become indispensable for fostering interdisciplinary integration, driving standardization, and enabling the creation of smart, resilient infrastructure.

Large Language Model (LLM) technology represents a breakthrough in the field of AI. Relying on unsupervised pre-training with massive text data, LLMs, such as ChatGPT and Qwen, have acquired extensive knowledge and reasoning abilities like human beings, thus making it possible to understand natural language and generate text close to human expression. Their excellent transfer learning capability enables them to strengthen knowledge and comprehension in specific domains through fine-tuning with a small amount of domain-specific data(Huy, 2019). Therefore, LLMs provide a possibility for more intelligent BIM-based design(Rong and Yu, 2025).

Building upon this potential, recent studies have explored multiple pathways for integrating LLMs into BIM-based design. These initiatives range from generating editable BIM models with complete semantics from natural language descriptions(Du et al., 2024; Hu, 2019), and automating the extraction of semantic information to structure complex model data(Han et al., 2025), to facilitating interactive visualization and display control through conversational interfaces(Lin et al., 2024). More critically, in the domain of design review, LLMs are increasingly leveraged to intelligently identify potential defects and evaluate regulatory compliance(Lin et al., 2023), where researchers have integrated ontological and deep learning models to establish robust automated checking frameworks for BIM-based design(Chen et al., 2024; He et al., 2025).

It can be seen that LLMs are integrating into and advancing BIM technology from a brand-new perspective(Paramesha et al., 2025). However, the civil engineering domain features high professional thresholds and relies heavily on extensive prior knowledge. A majority of these studies either directly employ pre-trained general large models or utilize Retrieval-Augmented Generation (RAG)(Lewis et al., 2020) for knowledge base supplementation to bridge domain-specific gaps. Yet, due to the lack of systematic LLM evaluation methods(Wang et al., 2024) and high-quality datasets, existing studies often exhibit suboptimal performance within the BIM-based design(Liu et al., 2023).

In response to these challenges, this study centers on BIM-based design and aims to establish a coherent framework for evaluating and developing domain-specific LLMs. To this end, we introduce an evaluation benchmark tailored to BIM-based design, together with quantitative indicators that enable structured assessment of LLM performance. We further propose a BIM-to-text conversion method and construct the corresponding BIM-derived datasets to support both evaluation and fine-tuning. Building on these components, a series of representative LLMs are examined, and a domain-adapted model, Qwen-BIM, is developed through targeted fine-tuning. The results highlight the value of the benchmark and datasets in guiding model selection and domain adaptation for BIM-based design. The remainder of this paper is organized as follows: Section 2 reviews related research; Section 3 presents the methodology, including dataset construction, capability evaluation, and fine-tuning strategies; Section 4 reports the experimental results; and Section 5 concludes the study.

## 2 Literature Review

### 2.1 BIM-based design

BIM has become a core enabling technology in the digital transformation of the construction industry and a key driver of paradigm shifts in architectural design(Wang et al., 2020). Compared with traditional CAD, BIM integrates geometric and semantic information into a unified digital model, enabling consistent representation of building geometry, components, materials, and lifecycle attributes(Volk et al., 2014). During the design stage, BIM enhances design accuracy and multidisciplinary coordination by supporting virtual pre-construction, early conflict identification, and optimization of design schemes(Azhar, 2011). Studies have shown that BIM-based design processes reduce design changes, shorten project cycles, and improve overall productivity(Gu and London, 2010; Succar, 2009).

Within current practice, BIM-based design mainly encompasses several foundational tasks. BIM modeling enables parametric description and geometric construction of building components, supporting early spatial reasoning and coordinated design development(Kim et al., 2024). Information extraction focuses on retrieving attributes, relationships, and constraints embedded in BIM models to support downstream analyses such as quantity takeoff and rule checking(Lin et al., 2016). Visualization and display control provide intuitive 3D interaction, including multi-view inspection and model exploration, which enhances design communication and decision-making(Rehman et al., 2023). Design review evaluates the integrity, compliance, and coordination quality of BIM models, helping identify conflicts and errors that would traditionally require extensive manual inspection(Ni et al., 2025).

From a technological evolution perspective, BIM has progressed from geometric modeling to information integration, and from isolated parametric tools to intelligent, collaborative environments(Saeed, 2011). The introduction of IFC standards addressed interoperability issues among heterogeneous software platforms(Pazlar and Turk, 2008), while cloud computing and WebBIM have enabled real-time, multi-regional collaborative design(Wong et al., 2014). The integration of VR/AR, IoT, and digital twin technologies has further endowed BIM with dynamic sensing and interactive simulation capabilities(Tan et al., 2024). Meanwhile, advances in visual computing and machine learning, including component recognition from point-cloud data and automated information completion, have pushed BIM toward more intelligent design workflows(Bloch and Sacks, 2018; Hu et al., 2024).

Despite these developments, BIM-based design still relies heavily on manual modeling and human-driven checking, and the level of intelligence achievable in practical applications remains limited(Liu et al., 2017). Enhancing automation and intelligence has therefore become a critical challenge in both BIM research and industry practice.

### 2.2 LLMs for BIM-based design

LLMs have emerged as a transformative force in artificial intelligence (AI), demonstrating exceptional semantic understanding and reasoning capabilities through large-scale pre-training. Their massive parameter scales facilitate strong transfer learning potential, enabling applications ranging from dialogue systems to knowledge-intensive reasoning. Furthermore, the emergent abilities of recent models(Wei et al., 2022a) have significantly accelerated the development of domain-specific LLMs in specialized disciplines.

In the realm of BIM-based design, the integration of LLMs is rapidly expanding, typified by several key research directions. Regarding BIM model generation, recent studies have progressed from conceptual text generation to precise model manipulation, such as utilizing LLMs to interpret natural language commands for modifying component attributes within Industry Foundation Classes (IFC) files directly(Besiktepe, 2025). In terms of information extraction, LLMs are employed to bridge the semantic gap, enabling the automated semantic enrichment of BIM models to infer missing properties required for structural safety analysis(Li and Zhang, 2024). For visualization and interaction control, conversational AI assistants have been developed to lower technical barriers, allowing users to intuitively query model information or control 3D viewpoints through real-time dialogue(Fernandes et al., 2024). Notably, in the critical domain of design review, LLMs are leveraging their semantic understanding to parse complex, unstructured regulatory texts into executable code or logic clauses for automated compliance checking(Zheng et al., 2022). Recent advancements leverage multimodal reasoning by combining visual recognition with textual interpretation to significantly enhance the robustness of automated compliance verification against building codes(Zhang et al., 2025).

However, despite these explorations, the application of LLMs in BIM-based design remains predominantly at the periphery rather than the core. Most current studies utilize LLMs for auxiliary detailed tasks, such as parsing specification texts into syntax trees, rather than directly entrusting them with core decision-making processes like defect identification and judgment(Shamshiri et al., 2024). For instance, while Prompt Engineering is widely applied, it often lacks the depth required to embed robust domain-specific capabilities, which leaves general models struggling to meet the high professional thresholds of engineering design(Chen et al., 2024). Consequently, in-depth application is hindered by systemic challenges. Wang et al.(Wang et al., 2024) pointed out that there is currently a lack of systematic LLM evaluation methods in BIM field, while Saka et al.(Saka et al., 2024) noted many challenges in LLM's applications, such as difficulty in data interoperability, high infrastructure costs, lack of domain-specific LLMs focusing on construction disciplines, and distrust of output.

Overall, while the introduction of LLMs into architectural design has opened new possibilities for semantic interaction and information representation in BIM-based workflows, current research predominantly relies on integrating general-purpose LLMs into BIM applications. These models are mostly limited to tasks such as information extraction or basic auxiliary functions, and they still struggle to support the complex semantic reasoning required in the design stage. The limitations arise from two major gaps. First, LLMs are primarily designed to process text or images and cannot directly interpret BIM models, which makes effective data conversion a necessary but unresolved step(O.A et al., 2024). Second, general LLMs lack pre-training on BIM and construction-related corpora, resulting in insufficient domain knowledge and reduced reasoning reliability in professional design scenarios(Wu et al., 2025). Consequently, existing BIM-related applications of LLMs remain constrained by the capabilities of general models, and the development of LLMs specifically adapted to BIM-based design tasks is still in its early stages and remains significantly underexplored.

**2.3 Development of domain-specific LLMs**

Developing domain-specific LLMs is essential to overcome the inherent limitations of general-purpose models in specialized engineering contexts. Although powerful, general LLMs struggle to accurately internalize deep domain knowledge and often require specific interventions, such as expanding context windows or fine-tuning, to bridge this gap(Dacheng Li, 2023; Shao et al., 2023). Furthermore, the inherent stochasticity of these models frequently leads to unstable outputs and hallucinations, rendering them ill-suited for engineering tasks that demand rigorous mathematical calculation, logical reasoning, and high precision(Zhang et al., 2024). Consequently, directly applying general LLMs to complex BIM-based design problems often yields suboptimal results, necessitating the development of models specifically adapted to the domain.

To construct such domain-specific models, fine-tuning based on pre-trained base models has become the prevailing approach, as training from scratch demands prohibitive computational resources and massive datasets(MathavRaj et al., 2024). Fine-tuning methods are generally categorized into Full Fine-Tuning and Parameter-Efficient Fine-Tuning (PEFT). While Full Fine-Tuning updates all model parameters to achieve deep adaptation, it is computationally expensive and requires extensive high-quality datasets(Pang et al., 2025). In contrast, PEFT freezes most pre-trained parameters and updates only a small subset, offering a balance between efficiency and performance(Singhal et al., 2023). Among PEFT techniques, Low-Rank Adaptation (LoRA)(Hu et al., 2021) has emerged as a representative method. By injecting trainable rank decomposition matrices into the Transformer architecture, LoRA significantly reduces the number of trainable parameters while maintaining the model's generative capabilities, making it a highly feasible solution for engineering applications with limited computational resources.

However, effective fine-tuning depends not only on efficient algorithms such as LoRA but also on the availability of high-quality domain datasets and evaluation benchmarks that are aligned with task requirements. Existing construction and BIM related corpora are typically designed for document understanding, regulation question answering, or general BIM information retrieval, and most of them contain only shallow question-answer (QA) pairs without supervision of intermediate reasoning steps(Aydın, 2022; Peng and Liu, 2023). As a result, they provide limited support for core design scenarios such as multi-step geometric reasoning, code-compliant parameter verification, or defect repair decision making, and they seldom include structured Question-Reasoning-Answer (QRA) data entries that are needed to represent complex BIM logic(Alnuzha and Bloch, 2025).

In parallel, widely used reference based text metrics such as BLEU and ROUGE mainly capture lexical similarity and have been shown to correlate weakly with factual correctness, reasoning validity, and constraint satisfaction, which are essential in engineering design tasks (Blagec et al., 2022; Evtikhiev et al., 2023). Consequently, these metrics cannot reliably determine whether a model output satisfies engineering constraints or complies with regulatory requirements. Without datasets and evaluation schemes tailored to domain reasoning and compliance checking, diagnosing model weaknesses and guiding targeted improvement for BIM based design tasks remains challenging.

Given these limitations, the development of domain-specific LLMs generally follows a clear and structured workflow. The process usually begins with constructing a task-oriented benchmark that reflects representative application scenarios and enables systematic evaluation of candidate base

models([Touvron et al., 2023](#)). Since engineering tasks require reasoning validity and compliance checking beyond lexical overlap, the benchmark is preferably paired with reasoning- and compliance-aware evaluation criteria. Based on these evaluation results, an appropriate base model is then selected by considering both its measured performance and practical deployment requirements. The benchmark results further reveal the strengths and weaknesses of different models, which in turn guide the creation of a targeted fine-tuning dataset enriched with domain knowledge and, when necessary, explicit reasoning supervision (e.g., QRA-style data with intermediate reasoning steps). The selected model is subsequently fine-tuned using parameter-efficient methods such as LoRA to better align it with domain-specific tasks([Hu et al., 2021](#)). In practice, this workflow can be iteratively repeated to refine the dataset and further improve model performance. A schematic representation of this workflow is provided in **Fig. 1**.

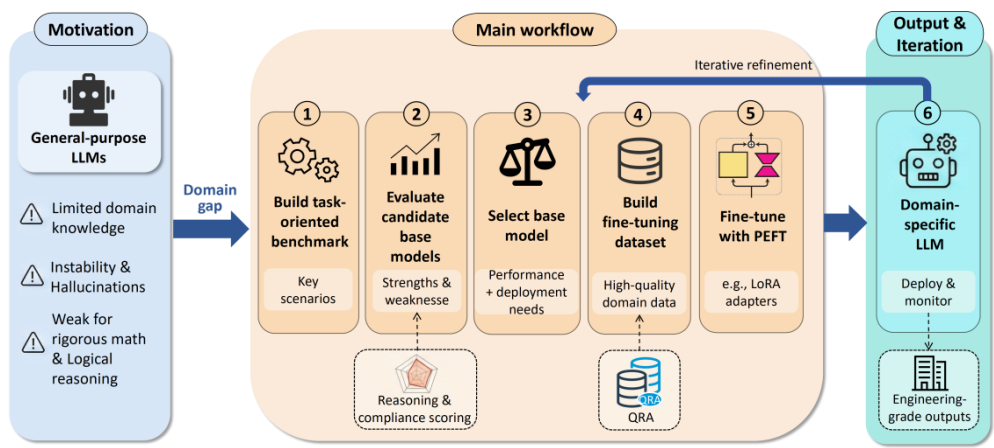

**Fig. 1.** General workflow for developing domain-specific LLMs

Overall, existing studies show that BIM-based design has formed a solid technical foundation across modeling, information extraction, visualization, and design review, yet its level of intelligence remains limited([Lin et al., 2023](#)). Although LLMs provide new possibilities for enhancing semantic understanding in BIM workflows, current applications mainly rely on prompting strategies or direct use of general-purpose models, confining them to relatively simple or auxiliary tasks([Zheng and Fischer, 2023](#)). These models lack embedded domain knowledge, cannot directly interpret BIM data, and operate without unified evaluation mechanisms. Meanwhile, the development of domain-specific LLMs depends on fine-tuning methods and high-quality datasets, but such resources for BIM-based design remain insufficient. As a result, existing LLM-based approaches are not well suited for reasoning-intensive or compliance-critical design tasks. Therefore, there is a clear need for specialized benchmarks, datasets, and fine-tuned models that can support the semantic and reasoning requirements of BIM-based design.

## 3 Methodology

**Fig. 2** illustrates the overall methodology. To develop a domain-specific LLM for BIM-based design, this study begins by establishing a task-oriented benchmark to specify representative design scenarios and the corresponding capability requirements for LLMs. Based on these task requirements, BIM data and associated documents are organized to construct the BIM-Question-Answer (BIM-QA) dataset and its reasoning-augmented counterpart (BIM-QRA), which provides both QA supervision and explicit reasoning supervision. For LLM evaluation, an evaluation protocol is designed that samples partitions from the BIM-QA dataset and forms standardized LLM inputs through prompt engineering (including the prompt template, evaluation date, and the question), after which model outputs are compared against reference answers and scored using a multi-dimensional indicator set that accounts for both lexical similarity and semantic consistency. On this basis, candidate general-purpose LLMs are systematically evaluated to analyze overall performance and task-wise strengths, and to compare open-source and closed-source models under a unified evaluation setting. The optimal foundation LLM is then selected, and LoRA-based fine-tuning is conducted using the reasoning supervision provided by BIM-QRA to obtain the domain-specific model, Qwen-BIM. This section details the research methodology in the sequence of benchmarks and indicators, dataset construction, LLM evaluation methods, and domain-specific LLM development.

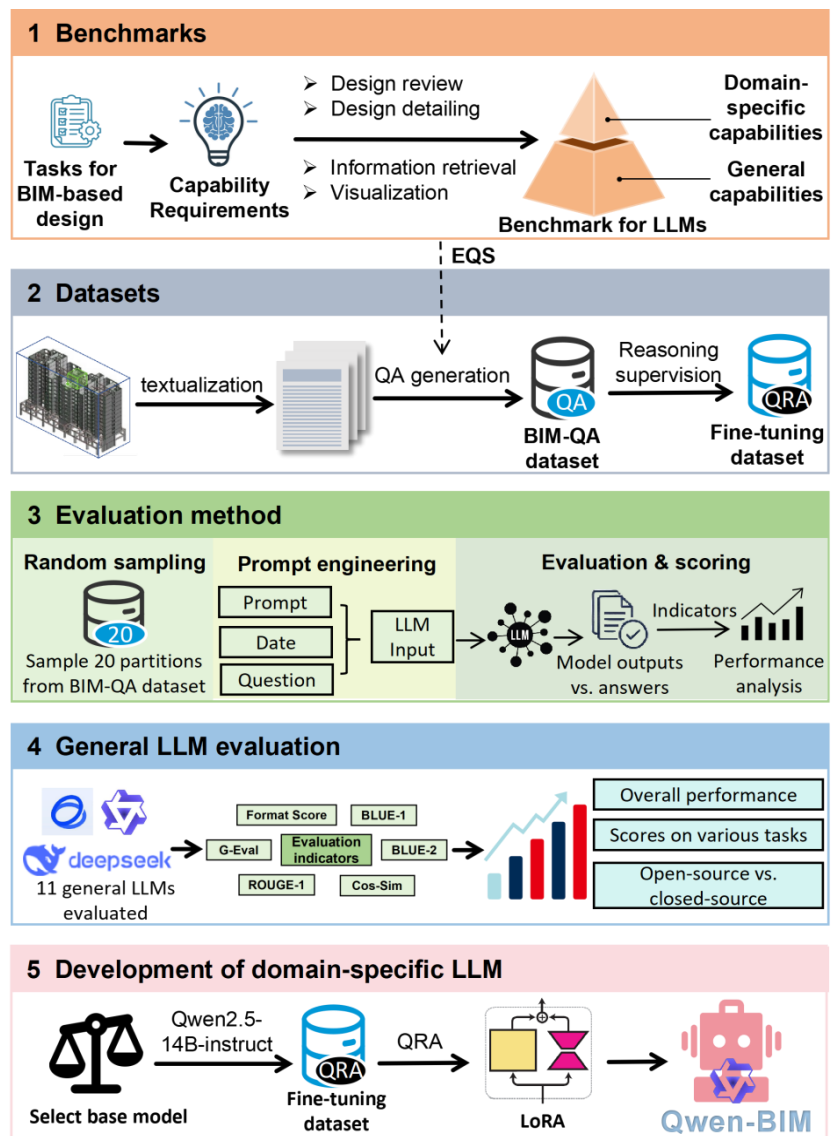

**Fig. 2.** Methodology

## 3.1 Benchmarks and Indicators

### 3.1.1 Capability requirements of LLMs

In this study, the capability requirements of LLMs are derived by analyzing typical tasks in BIM-based design workflows. We first examine practical design use cases and identify several representative task types, including information retrieval, visualization, various forms of design checking, and design detailing. For each task type, we enumerate the operations that an LLM is expected to perform and abstract the underlying abilities, which are summarized in **Table 1**.

For information retrieval, typical usage scenarios include querying the properties of specific components, locating elements that satisfy given conditions (for example, all fire doors on a floor), or asking for relevant clauses in design specifications. To support such tasks, LLMs need to understand the structure and semantics of BIM data, extract and filter information, perform basic statistics on quantities or ranges, and in some cases carry out simple calculations and lightweight reasoning to combine multiple constraints.

Visualization tasks mainly aim to transform BIM data into human-readable descriptions or summaries. Examples include generating textual descriptions of spaces, explaining the configuration of a system, or summarizing layout changes. In these scenarios, LLMs are expected to obtain and aggregate relevant information from BIM models, interpret the semantic meaning of components and their relationships, and organize the results into coherent narratives that assist designers in understanding the current design state.

Design review in terms of integrality, rationality, and compliance reflects three common viewpoints used in practical design checking. Integrality focuses on whether required components and parameters are present, such as whether all openings have associated fire doors or whether key attributes are missing. Rationality concerns whether existing parameter values are reasonable from an engineering perspective, for example whether thicknesses, spans, or clearances fall within plausible ranges. Compliance targets the consistency between BIM parameters and explicit specification requirements, such as verifying that fire resistance ratings or evacuation widths satisfy code constraints. These review tasks require LLMs not only to read and compare BIM parameters, but also to apply design common sense and specification knowledge, and in some cases to perform simple numerical calculations.

Design detailing represents a deeper level of assistance, where the model is expected to help refine design solutions after potential issues have been recognized. Typical examples include suggesting more appropriate parameter values within allowed ranges, recommending alternative component types, or describing adjustment strategies that better meet functional and regulatory requirements. Compared with the above tasks, design detailing places higher demands on multi-step reasoning, domain knowledge, and the interpretation of specification constraints.

Overall, the capability requirements extracted from these task types can be grouped into two categories. The first category consists of general capabilities, including information extraction, statistics, basic calculation, and reasoning. The second category consists of BIM domain-specific understanding capabilities, including the interpretation of BIM data, the use of design common sense, and the understanding of specification rules. Together, these capabilities provide a structured view of what LLMs need to support BIM-based design, and their necessity for each task type is summarized in **Table 1**.

Table 1 Capability requirements for LLMs in BIM-based design tasks

| Task type | | General capabilities | | | | BIM Domain-specific understanding capabilities | | |
|---|---|---|---|---|---|---|---|---|
| | | Extraction | Statistics | Calculation | Reasoning | BIM data | Common sense | Specification |
| Information retrieval | | √ | √ | ○ | ○ | √ | | |
| Visualization | | √ | √ | | √ | √ | | |
| Design review | Integrality | √ | √ | | | √ | | |
| | Rationality | √ | √ | | √ | √ | √ | |
| | Compliance | √ | √ | ○ | √ | √ | | √ |
| Design detailing | | | | ○ | √ | √ | √ | √ |

√ indicates that the ability is necessary, while ○ indicates that the ability may be needed.

**3.1.2 LLM evaluation question set (EQS)**

In line with the capability requirements defined in **Section 3.1.1**, an LLM EQS is constructed to quantitatively assess how well different models support BIM-based design tasks. The EQS covers both general capabilities and BIM domain-specific understanding capabilities, and each capability is mapped to several evaluation aspects such as accuracy of information extraction, correctness of numerical calculation, consistency of geometric reasoning, and adequacy of design-related judgments. As summarized in **Table 2** and **Table 3**, the EQS consists of 22 parameterized question templates. Placeholders marked with "[xxx]" are instantiated according to the actual BIM model or design scenario so that multiple concrete questions can be generated from each template. In general, **Table 2** contains question templates for general capabilities including data extraction, statistics, calculation, and reasoning, while **Table 3** contains templates that embed domain knowledge of BIM design and specifications to test integrality, rationality, compliance related judgments and corresponding adjustment suggestions.

Building on this design, **Table 2** presents the EQS corresponding to the general capabilities. Each type of question is set with different levels of difficulty to test the LLM's capability. For data extraction questions, Level 1 only requires LLMs to extract certain specific data, while Level 2 demands extracting all components within a specific category, which is relatively more challenging. In statistics questions, Level 1 merely involves counting, whereas Level 2 requires classification and statistics based on specific attribute. Regarding calculation questions, Level 1 is simple length calculation; Level 2 involves calculating and determining vector directions; and Level 3 entails 2D area calculation. Finally, for reasoning questions, Level 1 focuses on the inference of wall orientation; Level 2 assesses the understanding of design common sense and BIM data to infer building floor heights; and Level 3 pertains to the simplified version of collision detection, i.e., collision detection in 2D planes (it is worth noting that early tests revealed that LLMs struggle with data calculations in 3D spaces, such as collision detection, so this study does not include questions involving 3D calculations). Solving Level 3 questions requires LLMs to fully understand the geometric characteristics of "wall" data, then comprehensively analyze parameters such as the position and size of the walls.

Table 1 LLM EQS for general capabilities

| Capability | Level | Question templates |
|---|---|---|
| Data Extraction | Level 1 | What is the ID of [the first wall] in the model, and what is [the wall thickness]? |
| | Level 2 | How many [walls] are included in this model? Please list the ID, start coordinate, and end coordinate of each one. |
| Statistics | Level 1 | How many [beam] components are there in the model? |
| | Level 2 | How many types of wall thicknesses are there in the model? Please list the IDs of all walls included in each wall thickness. |
| Calculation | Level 1 | Based on the coordinates of the start and end positions of the [second] wall, calculate the length of the wall in millimeters. Compare the calculation result with the given length data to determine whether the given data is accurate. |
| | Level 2 | Please calculate the direction of each wall according to the start and end coordinates. Describe the direction using "along the positive x-axis", "along the negative x-axis", "along the positive y-axis" and "along the negative y-axis". |
| | Level 3 | Please calculate the planar area of all rectangular slabs in this model, with the calculation results in square meters. In your answer, please first list the ID of each slab, indicate whether it is rectangular, and if so, print its area. After listing the results of all slabs, provide the ID and area of the largest slab among them. |
| Reasoning | Level 1 | Are the planar orientations of the [second] wall and the [third] wall parallel to each other, perpendicular to each other, or neither parallel nor perpendicular? |
| | Level 2 | What is most likely the floor height of the building? Please briefly explain the basis for your judgment. |
| | Level 3 | Based on the analysis and judgment of the planar geometry range of walls, do the [fifth] wall and the [third] wall have any overlapping? |

**Table 3** presents the EQS for BIM domain-specific understanding capabilities. The questions are organized around typical BIM-based design scenarios, because these capabilities are difficult to assess through single, isolated operations. The task types include design review from the perspectives of integrity, rationality, and compliance, as well as design detailing. For each task type, Case 1 and Case 2 describe two representative scenarios with the same task objective but different attributes, rules, or contextual information. This design allows the evaluation to examine whether the model can generalize its domain understanding across similar design situations.

For integrity related checks, the design review questions ask the model to determine whether components have required attributes such as construction numbers or fire resistance ratings and to list the IDs of components with missing data. The associated integrity detailing questions then provide naming or parameter rules and require the model to complete the missing attributes accordingly, thereby testing whether it can correctly interpret and apply domain-specific constraints.

For rationality related tasks, the design review questions focus on wall thickness and wall z-coordinate attributes, asking the model to detect suspicious values, for example excessively large or small thicknesses or inconsistent z-coordinates at the start and end points of a wall. Prompts such as "According to [the thickness range of walls in general residential buildings]" and "According to [general building common sense]" explicitly guide the model to reason on the basis of design common sense. The corresponding rationality detailing questions then require the model to propose reasonable replacement values, evaluating whether its suggested modifications are consistent with typical engineering practice.

For compliance oriented tasks, the design review questions cover both user defined rules and specification based provisions, such as checking whether construction numbers satisfy given naming rules or whether fire resistance ratings meet minimum code requirements. The related compliance detailing questions ask the model to modify non-compliant data so that it satisfies the specified rules or clauses. When instantiating these templates, different ways of referring to components, for example by serial number or by explicit IDs, are adopted to enrich the scenarios and to further probe the logical reasoning and domain understanding capabilities of LLMs.

**Table 3** LLM EQS for BIM Domain-specific understanding capabilities

| Task type | Case | Question templates |
| --- | --- | --- |
| Design review - Integrality | Case 1&2 | Please check if all components in the model have been provided with [construction number / fire resistance]. If there are components missing this attribute, please list their IDs. |
| Design review - Rationality | Case 1 | According to [the thickness range of walls in general residential buildings], infer whether there are any suspicious [wall thickness data]. If so, please list the IDs of the suspicious data and the reasons for suspicion. |
|  | Case 2 | According to [general building common sense], [based on the z-coordinate of all walls], infer whether there are any suspicious [z-coordinate]. If so, please list the IDs of the suspicious data and the reasons for suspicion. |
| Design review - Compliance | Case 1 | Here are naming rules for component construction numbers: A component's construction number consists only of one initial letter followed by numbers (e.g., Q20). The construction number of a wall must start with "Q", that of a beam must start with "L", and that of a slab must start with "B". Please check whether the construction numbers of all components in the model comply with these requirements. If a component's construction number is missing, it will not be checked. Please list the IDs of the components whose construction numbers do not comply with the naming rules among the components that need to be checked. |
|  | Case 2 | Here is a design rule: The fire resistance of structural beams must be greater than or equal to 2 hours. Please check each structural beam in the model one by one to test if they meet this requirement. If a beam lacks this parameter, it will not be checked. Please list the IDs of the beam components that do not comply with the fire resistance requirements among the beam components that need to be checked. |
| Design detailing - | Case 1 | Here are naming rules for component construction numbers: [xxx]. When completing the construction number, determine |

| | Integrality | | the letter according to the component type, and then uniformly use "00" for the numbers. Please complete the missing attribute for [the first wall / the beam with ID 215431] in accordance with the above rules. |
|---|---|---|---|
| | | Case 2 | Here is a design rule: [xxx]. Based on the existing fire resistance data of structural beams and the above design rule, please provide a simple reason to infer and complete this data for [the last beam] that is missing this attribute. |
| | Design detailing - Rationality | Case 1 | According to the wall thickness range in general residential buildings, the wall thickness of [the third wall] is unreasonable. Please propose a reasonable data to modify it based on design common sense and other normal wall thickness data. |
| | | Case 2 | According to general building model common sense, the z-coordinates of the start and end points of the same wall should be the same and consistent with the z-coordinate data of other walls on the same floor. Now, the z-coordinate attribute of [the second wall] is unreasonable; please propose a reasonable data to modify the z-coordinate. |
| | Design detailing - Compliance | Case 1 | Here are naming rules for component construction numbers: [xxx]. Now, the construction number of the components with IDs [234561, 289102] is non-compliant. Please provide a compliant data to modify this attribute. |
| | | Case 2 | Here is a design rule: [xxx]. The fire resistance of the beams with IDs [215431, 102351] is non-compliant. Please provide a compliant data to modify this attribute. |

**3.1.3 Evaluation indicators**

To quantify the similarity between the output results of LLMs and the preset answers, evaluation indicators based on text similarity and semantic similarity are constructed. Text similarity is obtained by finding morphemes, words, and other elements that appear simultaneously in the output text and the answer text, and counting the length and quantity of such co-occurring elements mainly to determine the similarity in text. Semantic similarity measures how closely related two pieces of text are in terms of their meaning, which is no longer limited to the reproduction of identical text, often used in NLP applications. In this study, semantic similarity is judged based on deep learning networks or LLMs. The final evaluation indicators are shown in **Fig. 3**.

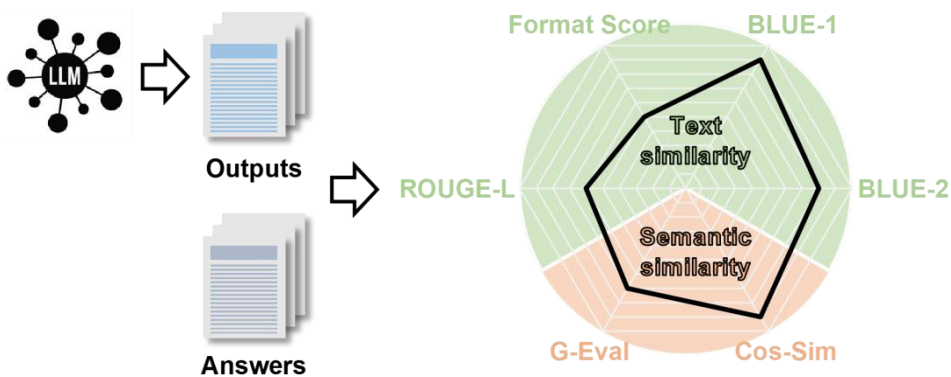

**Fig. 3.** LLM quantitative evaluation indicators

(1) Text similarity related indicators

Text similarity indicators measure the lexical and structural overlap between the model output and the reference answer, and mainly rely on exact or near-exact matching of tokens and subsequences. They are easy to compute automatically but are relatively sensitive to wording and formatting. In this study, text similarity is evaluated using a format score, BLEU-1 and BLEU-2, and ROUGE-L. The specific definitions of these indicators are given below.

1) Format Score. Although system prompts are used to specify the answer format, LLMs may still return outputs that do not follow the required structure. The format score is therefore introduced to measure instruction-following. It takes only two values, 1 for strict compliance and 0 for non-compliance. Once the output does not meet the format requirements, the entire text is still used for subsequent indicators, which may lower their scores even if part of the content is correct.

2) BLEU-1 and BLEU-2. The BLEU (Bilingual Evaluation Understudy) metric(Papineni et al., 2002) was originally proposed for machine translation and measures the proportion of n-grams in the candidate output that also appear in the reference. In preliminary experiments, higher order BLEU scores such as BLEU-4 were close to zero because of the diversity of answers, so this study adopts BLEU-1 and BLEU-2 only. It should be noted that, although BLEU is widely used, recent studies have reported weak correlation with human judgments in LLM-related tasks(Evtikhiev et al., 2022), and its applicability in this context is therefore examined with caution.

3) ROUGE-L. ROUGE (Recall-Oriented Understudy for Gisting Evaluation) (Lin, 2004) is a series of evaluation indicators, initially used to assess the performance of text summarization and abstract generation. Among them, the ROUGE-L indicator uses the Longest Common Subsequence (LCS) to describe the similarity in sentence structure and word order. Let the answer be denoted as A and the actual output as B. Then the ROUGE-L indicator is given as follow:

$$R_{lcs}=LCS(A,B)/Length(A)$$

$$P_{lcs}=LCS(A,B)/Length(B)$$

$$F_{lcs}=(1+\beta^2)R_{lcs}P_{lcs}/(R_{lcs}+\beta^2 P_{lcs})$$

Here, LCS(A, B) refers to the LCS of A and B; Length(A) represents the length of A, calculated by characters. $F_{lcs}$ is the ROUGE-L indicator, and β is an adjustable parameter.

Compared with the BLEU metric, ROUGE-L can capture the overall similarity of long texts, so it is more suitable for evaluating the similarity between ultra-long texts or texts with significant length differences. In this study, the outputs of LLMs are random, which may lead to a huge difference in length between the output and the answer. Therefore, this metric is adopted to reduce the impact of such situations on the evaluation results.

(2) Semantic similarity related indicators

Semantic similarity indicators focus on whether the output preserves the intended meaning of the reference, even when the wording or sentence structure differs, and are therefore less sensitive to surface-level variations. In this study, semantic similarity is assessed by cosine similarity between embedding vectors and by the LLM-based G-Eval score. The specific settings of these indicators are described as follows.

1) Cosine Similarity (Cos-Sim). Cosine similarity is a commonly used indicator for evaluating semantic similarity(Chandrasekaran and Mago, 2020). Firstly, the actual output and answer are converted into high-dimensional vector representations through word embedding, and then the cosine value of the angle between them in the high-dimensional space is calculated. This indicator can reflect the semantic similarity of two texts, such as tendency, yes-no semantics, etc. In this study, the Embedding-3 provided by ChatGLM is used to map the text into a 1024-dimensional vector space.

2) G-Eval. In 2023, Liu et al.(Liu et al., 2024) proposed DeepEval, an open source framework that uses LLMs to evaluate LLM outputs. Benefiting from their natural language understanding ability, LLMs can accept evaluation instructions and scoring rules in natural language and return both scores and explanations. In this study, the GLM-4-plus model provided by ChatGLM is used to compute the G-Eval score. The model output and the reference answer are fed to GLM-4-plus together with an evaluation prompt, and the model returns a score between 0 and 1 and a brief rationale. To improve the stability of the scores, the temperature of GLM-4-plus is set to 0.

To guide G-Eval in giving reasonable scores, this study iteratively optimized the evaluation prompt, and the final prompt is as follows:

"*It is expected that the output is as semantically consistent with the answers as possible. Scores are given based on whether the semantic information of the actual output is complete and accurate. The following rules must be observed when scoring:*

*1. Differences in expression methods will not lead to deductions, including but not limited to symbol difference, units (such as 'millimeter' and 'mm'), the order of listed information (including but not limited to the listing of IDs and categories), missing or additional function words, etc.*

*2. Differences in reasons and causes will not lead to deductions. No deductions will be made regardless of whether the output or answers contain reasons and causes, and regardless of whether the reasons in the actual output are consistent with those in the answers.*

*3. The accuracy of judgmental results and data accuracy account for 90% of the total score. If there is an error in 'yes-no judgments' such as yes or no, vertical or parallel, presence or absence, positive or negative, all scores for this part should be deducted directly. For data information, classification considerations are required: For data obtained through calculations such as length and area, there may be rounding errors, and an error within 3% is considered accurate; For data such as serial numbers, IDs, and component attributes, complete consistency is required to be considered accurate; If the answer provides multiple acceptable values connected by 'or', it is considered accurate as long as one piece of data information in the actual output matches. If there are multiple pieces of information, the score is based on the percentage of accurate data information in the total number of information pieces. For example, if the judgmental information is accurate but half of the data information is accurate, half of the scores will be deducted; if 80% is accurate, 20% of the scores will be deducted; if completely correct, no scores will be deducted.*

*4. If the length of the actual input is more than 3 times that of the answer, 10% of the total score will be deducted.*

*5. The total score is obtained by deducting scores through the above 4 steps, starting from the full score.*"

This prompt specifies the scoring standards in detail, clarifies the situations where no scores are deducted, provides methods and examples for score calculation, and can ensure the stability and reliability of LLM scoring as much as possible.

### 3.2 Dataset Construction
### 3.2.1 Development of the BIM-QA dataset

BIM data is inherently complex and highly structured, regardless of whether it is stored in proprietary formats (such as .rvt files in Revit) or in universal standards like IFC. Such representations exhibit low human readability and fall outside the natural-language domain that LLMs can directly interpret. Wang et al.(Wang et al., 2024) pointed out that the complex expression of the IFC format makes it difficult to extract independent primitives from it. So, this study proposes a method for semi-structured BIM data textualization. On the one hand, natural language expressions are used as much as possible to make them closer to human expressions of BIM, thereby reducing the difficulty for LLMs to understand. On the other hand, it retains advantages of structured expressions, featuring strong scalability and being applicable to describing various components and data types. On this basis, as shown in **Fig. 4**, BIM models are first partitioned and textualized into BIM inquiry text, which serves as the common input for QA construction. For general tasks, questions are generated from the inquiry text to cover extraction, statistics, calculation, and reasoning, and the corresponding answers (ground truth) are derived to form general-task QA pairs. For domain tasks, artificial defects are further injected into the BIM inquiry text to obtain defect-augmented inquiry text, based on which domain-task QA pairs are generated. By aggregating QA pairs from both general and domain tasks, the resulting evaluation dataset is referred to as the BIM-QA dataset, which contains BIM textual descriptions paired with their corresponding questions and answers. Among them, both the BIM models and injected defects are stored in the dataset in textual form.

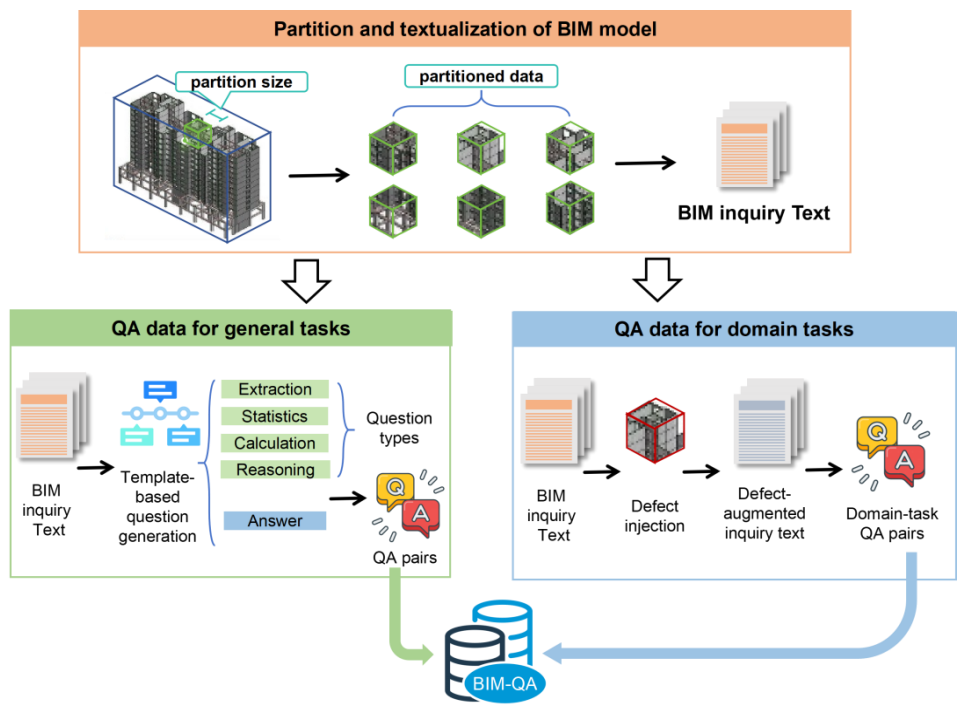

**Fig. 4.** Construction of evaluation dataset for LLMs in BIM-based design domain

**Fig. 5** illustrates the textualization process for BIM data. As shown in the figure, the Header Section provides an overall description of the BIM model, specifying the model type and the included component categories. For example: "This is part of a single-story building model, including three types of components: shear walls, structural beams, and slabs."

Following this, each component category is introduced through a Component Header Section, which explains the category name and the key information required for describing its instances. Taking shear walls as an example, the Component Header Section specifies the semantic meaning of attributes such as starting and ending coordinates, thickness, height, and length, as well as their units.

Based on the category header, each individual component is then described in the Component Data Section using clear, natural-language expressions. For instance: "The ID of the 1st wall is 342693. Its starting point coordinate is (-1897, -5891, 0), and the ending point coordinate is (-1897, -3191, 0). The wall thickness is 200 mm, the wall height is 2900 mm, the wall length is 2700 mm, and its construction number is Q23."

For components with more complex geometric attributes, such as slabs with polygonal or L-shaped outlines, the Component Data Section also adopts descriptive natural-language forms to convey the spatial configuration. As illustrated in **Fig. 5**, the slab shape is expressed through ordered corner-point coordinates along its upper surface, enabling accurate textual representation of irregular geometries.

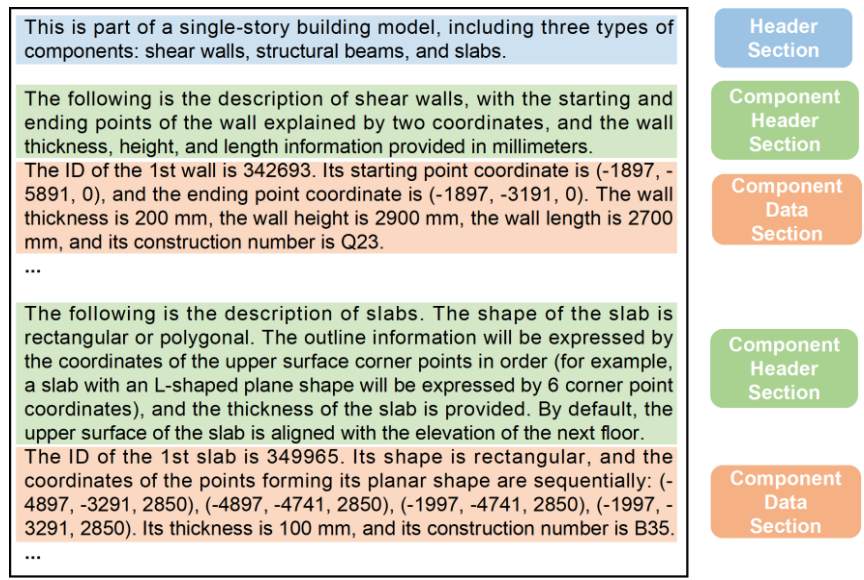

**Fig. 5.** Example of BIM data textualization

The EQS provides a question dataset for LLM evaluation, but it still needs to be combined with BIM data as input to the model. Moreover, the BIM data must cover representative design tasks and scenarios, whereas in engineering practice it is difficult to obtain sufficient real-world defective data. Therefore, this study proposes a method for automatically generating defect data based on complete BIM models, constructing BIM-QA dataset. The method includes the following three steps.

(1) Model inspection: Inspect the integrity, rationality, and compliance of the given BIM models. Revise and supplement incorrect parameters to ensure that the BIM data meets the design requirements.

(2) Model partitioning and textualization: According to the input length limit of LLMs and their processing capability for long texts, partition the BIM data of all components. A block of partitioned data is obtained by packaging a limited number of components within a certain adjacent spatial range. On this basis, textualize it using the above method, and this text is the model data in BIM-QA dataset.

(3) BIM inquiry text generation: For general tasks, any BIM model text can be used as an inquiry input. For domain specific tasks, the BIM model must first be augmented with defect data. According to the specific questions in the EQS, randomly select components in the partitions and modify their parameters to generate defects, ensuring that at least one defect is generated for each type of question. For integrity, randomly erase the construction number or fire resistance of components in the original data. For rationality, randomly add the following three types of defects: a) Modify the wall thickness to be outside the range of 20-600mm; b) Modify the Z-coordinate of the wall's start or end point (e.g., increase by 120mm); c) Modify the Z-coordinates of both the start and end points of the wall simultaneously to make it move integrally compared to other walls. For compliance, randomly add the following defects: a) Add extra letters or symbols to the construction number, or delete letters, such as modifying the compliant "Q20" to "WQ20", "Q-20", or "20"; b) Modify the fire resistance to the non-compliant "1 hour". Above texts with defects are the BIM inquiry text in BIM-QA dataset. And the EQS questions are the question data in BIM-QA dataset.

(4) Answer generation: Generate answers (ground truth) for each question in (3) according to the original BIM data. And finally, the BIM inquiry text, questions, and answers are combined into BIM-QA dataset, which serve as the final BIM evaluation dataset.

**3.2.2 Fine-tuning dataset with reasoning processes**

To train and fine-tune domain-specific LLMs, it is necessary to teach them knowledge reasoning methods in the engineering field. Answers of the BIM-QA dataset constructed in **Section 3.2.1** often only include results without reasoning processes. Therefore, a QRA dataset needs to be built to assist in the fine-tuning of LLMs. In particular, the QA process obtained from LLM evaluation in **Section 3.3** implicitly contains the reasoning processes of LLMs (LLMs with reasoning ability always reason first before outputting results), which can be quickly converted into QRA triad. This makes it the most convenient data for fine-tuning. LLMs with deep reasoning capabilities such as DeepSeek-R1 generate quite detailed reasoning processes when answering questions. Such accurate data will help LLMs learn reasoning approaches, enhance their domain-specific capabilities, and optimize the effectiveness of fine-tuning.

The method to acquire fine-tuning data is as follows. Firstly, an additional 20 partition data are fed to four LLMs (GLM-4-plus, Qwen-plus, DeepSeek-R1, and QwQ-32b), which can generate detailed reasoning processes. Then, from the above record data of LLM evaluation, the outputs with a format score of 1 and a G-Eval indicator of 0.8 or higher are screened to ensure that the outputs meet the format requirements and accuracy. Meanwhile, to compare the impact of reasoning data on LLM fine-tuning, QA data and QRA data are saved separately. Finally, through the above methods, a total of 3,493 pairs of data are obtained, including 2,129 pieces of QA dataset and 1,364 pieces of QRA dataset. All selected samples are further manually checked to verify the correctness of both the answers and the reasoning processes. The example of QA data and QRA data is shown in **Fig. 6**.

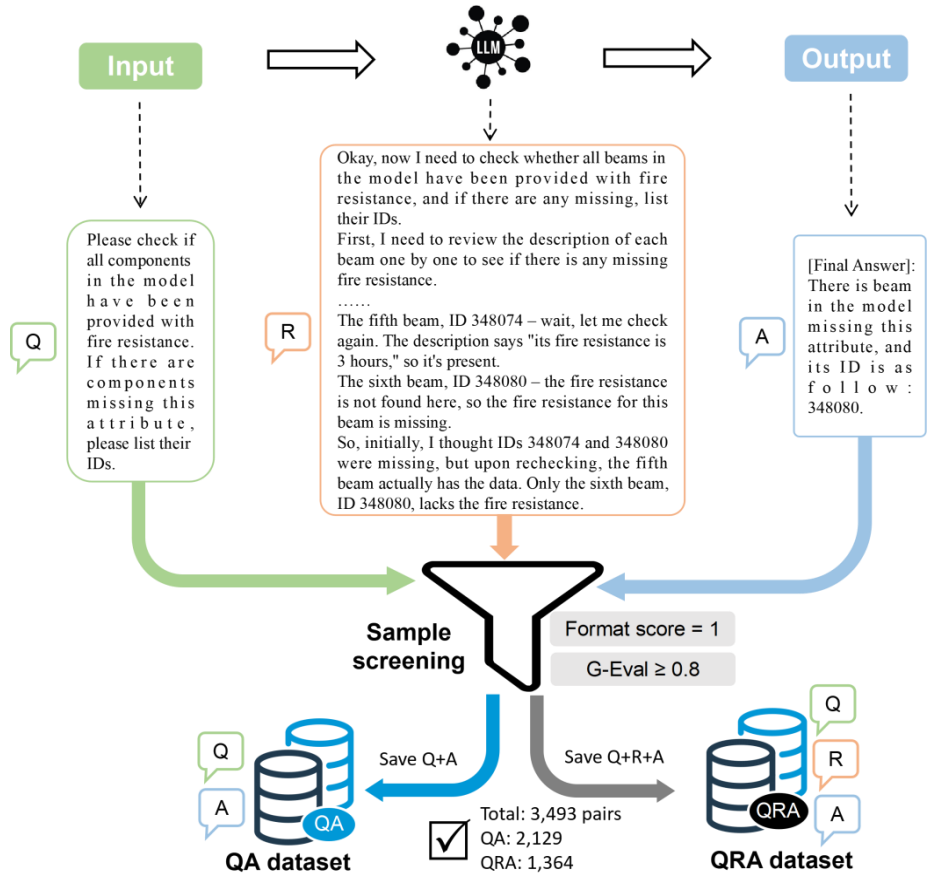

**Fig. 6.** Example of QA data and QRA data

### 3.3 LLM Evaluation Methods

As illustrated in **Fig. 7**, the evaluation of LLMs in this study follows an integrated method that combines prompt engineering with quantitative similarity-based scoring. The prompt engineering stage reconstructs each test sample from the BIM-QA dataset into a standardized LLM input, consisting of a system prompt, BIM inquiry text, and task-specific questions, thereby enforcing consistent output formats and mitigating response randomness. Based on the generated outputs, a multi-metric evaluation scheme is employed, including text-similarity indicators (Format Score, BLEU-1/2, and ROUGE-L) and semantic-similarity indicators (Cos-Sim and G-Eval), where only G-Eval is rated by an LLM. Using this unified evaluation method, controlled experiments are conducted to benchmark general-purpose LLMs and to identify the most suitable foundation model for subsequent domain-specific fine-tuning.

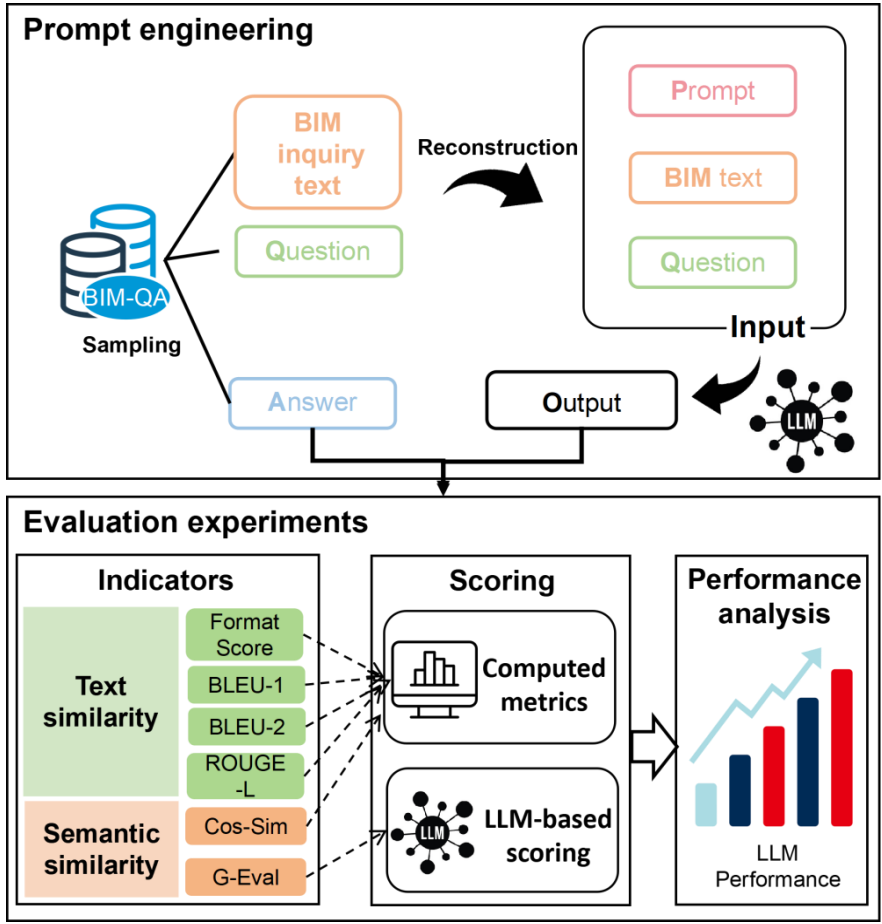

**Fig. 7.** Automatic LLM evaluation method

### 3.3.1 Prompts for evaluation

As shown in **Fig. 8**, based on the prompt learning, the evaluation input consists of four parts: system prompt + textualized defect data + question + example prompt.

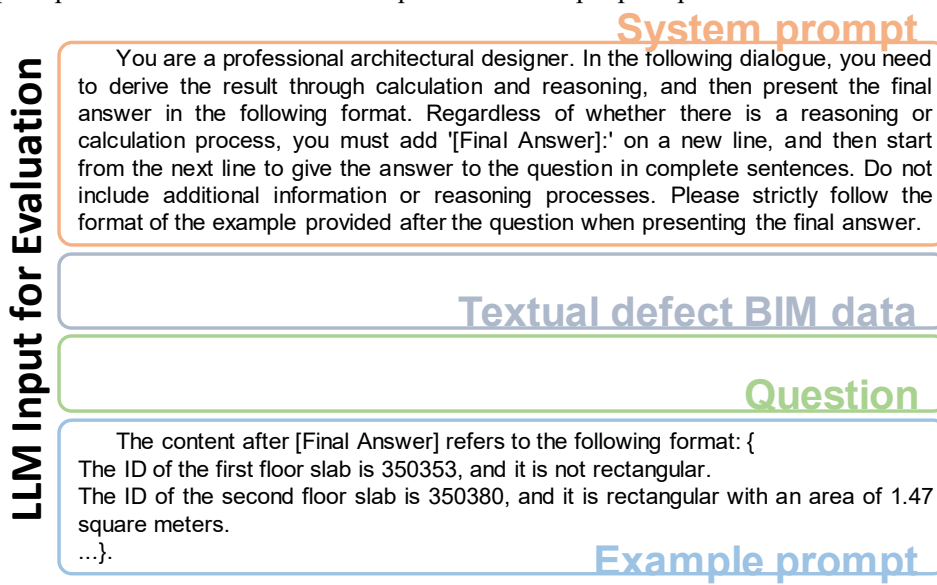

**Fig. 8.** Automatic LLM evaluation method

（1）System prompt

In prompt learning, system prompt is usually used to set global rules for LLMs. Research (Bsharat et al., 2023) shows that by setting a "role-play" in the system prompt, LLMs can output content more in line with the role identity and domain scene. The process of prompt optimization indicates that it is necessary to further guide LLMs to adopt the "chain-of-thought" approach for step-by-step reasoning to improve the accuracy of answers. In addition, it is also necessary to describe the format requirements for the outputs in clear and unambiguous words. Setting format requirements is, on the one hand, to test the large model's ability to follow instructions; on the other hand, the standardized format facilitates the subsequent calculation process of indicators to quantify the evaluation results. The output of LLMs is random, and the reasoning processes vary greatly. Therefore, this study only considers the accuracy of the "final answer" when evaluating the output of LLMs. Finally, the system prompt also needs to clearly indicate the example content after the question, so as to reduce the impact of the randomness of different models on the evaluation results.

After experimental optimization, the system prompt set in this study is as follows: "*You are a professional architectural designer. In the following dialogue, you need to derive the result through calculation and reasoning, and then present the final answer in the following format. Regardless of whether there is a reasoning or calculation process, you must add '[Final Answer]:' on a new line, and then start from the next line to give the answer to the question in complete sentences. Do not include additional information or reasoning processes. Please strictly follow the format of the example provided after the question when presenting the final answer.*"

（2）Textual defect BIM data and question

According to the EQS, each partition BIM data (including defect) contains 22 questions. Before each question is posed, the corresponding textual defect BIM data needs to be added as part of the prompt to provide the specific context and relevant data for the question. The textual defect data itself is self-explanatory, so no additional supplementary explanations are required.

（3）Example prompt

To further standardize the format of the "final answer", few-shot prompts with answer examples(Wei et al., 2022b) are set up, aiming to obtain outputs that better meet users' expectations. In this study, the Example prompt is added after the question and adopts the following format: *The content after [Final Answer] refers to the following format: example text*. The content in is an example corresponding to the specific question. The method for generating examples is to randomly select a partition of BIM data, manually hide parts of the answer to each question, and replace them with "...", thereby obtaining example texts for all questions. For instance, for the area calculation question in calculation capability level 3, the example text is:

"*The ID of the first floor slab is 350353, and it is not rectangular.
The ID of the second floor slab is 350380, and it is rectangular with an area of 1.47 square meters.
...
Among these, the ID of the rectangular floor slab with the largest area is ..., and its area is ... $m^2$.*"
Here, more listed items as well as the specific ID and area information of the largest floor slab are hidden. Another type of hidden information is related to questions that require explanations of reasons. For example, the answer example for Case 1 of rationality identification is:

"*There are some suspicious wall thickness data in this part of the model. The thickness of the first wall is suspicious; its ID is 342679, and its thickness is 20mm. Due to ..., this data may be incorrect.*"
The specific explanation of the reason is hidden here to prevent the data in the example from affecting the reasoning of LLM and causing hallucinations.

### 3.3.2 Evaluation experiment

In this study, the automatic LLM evaluation method consists of three steps: 1) Randomly sample 20 partition data from the generated BIM-QA dataset; 2) Based on the prompt learning method, supplement system prompt, answer requirements, defect model data, and example information to form a complete LLM input; 3) Conduct batch testing by remotely calling the LLM's API, compare the obtained outputs with the answers, and calculate quantitative evaluation indicators to analyze the LLM's capabilities.

To compare the capabilities of general LLMs with different architectures and developers on BIM based design tasks, this study selected 11 general LLMs (as shown in **Table 4**), including the ChatGLM, Qwen, and DeepSeek series. In the evaluation experiment, 20 partition data were extracted from the BIM-QA dataset, and the above-mentioned LLMs were tested one by one to calculate quantitative indicators. In order to analyze the capability gaps among LLMs and select available base LLMs for fine-tuning , this study designed three categories of experiments, and conducted analyses on the results of overall capability of general LLMs, capability on different types of questions, and comparisons between open-source LLMs.

Table 4 General LLMs used in evaluation

| Series | Version | Parameter count (billion) | Evaluation date | Reasoning |
|---|---|---|---|---|
| ChatGLM | GLM-4-plus | - | 2025-03-14 | × |
| | GLM-4-0520 | - | 2025-03-14 | × |
| | GLM-4-flash | - | 2025-03-14 | × |
| Qwen | Qwen-max | - | 2025-03-15 | × |
| | Qwen-plus | - | 2025-03-15 | × |
| | Qwen-turbo | - | 2025-03-15 | × |
| | Qwen2.5-72b-instruct | 72 | 2025-03-16 | × |
| | Qwen2.5-32b-instruct | 32 | 2025-03-16 | × |
| | Qwen2.5-14b-instruct | 14 | 2025-03-16 | × |
| | Qwen2.5-7b-instruct | 7 | 2025-03-16 | × |
| | QwQ-32b | 32 | 2025-03-17 | √ |
| | QwQ-plus | - | 2025-03-17 | √ |
| DeepSeek | DeepSeek-V3(Original) | 671 (MoE, activation: 37) | 2025-03-18 | × |
| | DeepSeek-R1 | 671 (MoE, activation: 37) | 2025-03-18 | √ |

*Parameter quantity '-' represents a closed source model with non-public parameter quantities*

For the evaluation of overall capability of general LLMs, six closed-source non-reasoning LLMs from the ChatGLM and Qwen series were selected. This approach not only assesses the comprehensive capabilities of LLMs for target tasks, but also identifies the discriminative power of different indicators for LLMs, so as to further optimize evaluation system. On this basis, the winners among the closed-source models were compared with reasoning models to investigate the importance of reasoning capabilities in enabling LLMs to solve various problems, and to contrast the impacts of different parameter sizes on model performance. Regarding the capability on different questions, the experiments specifically selected two LLMs with different parameter sizes for comparison, aiming to analyze the problem-solving abilities of LLMs for both general problems and domain-specific problems. Finally, to select the most appropriate base LLM for fine-tuning, open-source LLMs from the Qwen series with different parameter sizes were evaluated and compared. The final base LLM was determined by integrating the experimental conditions.

**3.4 Development of the domain-specific LLM**

The strategies for fine-tuning and evaluation are as follows. Firstly, deploy the LLM locally, then use the LoRA method for fine-tuning based on the dataset constructed in the previous section. During fine-tuning, the GPU memory usage is controlled within 96G (2 Nvidia RTX A6000). The fine-tuning framework adopts llama-factory(Zheng et al., 2024), with parameters visualized through WebUI. After fine-tuning, apply the evaluation method proposed in Section 3.1 to test LLM's capability by another 20 partitioned data from the BIM-QA dataset again. The key fine-tuning parameters are shown in **Table 2**, and the remaining parameters use default settings.

It should be noted that the fine-tuning dataset used by the llama-factory framework requires an additional "Instruction" to be specified, which is similar to the prompt given to the LLM during fine-tuning. Therefore, in this study, the system prompt is input as the "Instruction", and other model text, questions, examples prompt, etc., are input as "Question" data during fine-tuning.

Table 2 Key parameters of fine-tuning

| Name | Value | Name | Value |
|---|---|---|---|
| finetuning_type | lora | lora_alpha | 32 |
| lora_dropout | 0 | lora_rank | 8 |
| cutoff_len | 3072 | plot_loss | true |
| gradient_accumulation_steps | 8 | learning_rate | 5.0e-5 |
| num_train_epochs | 2 or 3 | warmup_steps | 5 |

To explore the impact of reasoning data on LLM fine-tuning, comparative experiments are conducted using datasets with different mixtures of QA dataset and QRA dataset. As shown in **Table 3**, datasets with different mixing proportion of QA dataset and QRA dataset are used to compare the impact of reasoning data on model performance. In particular, since Group D is fine-tuned only with the QRA dataset, which has a smaller amount of training data, the "num_train_epochs" is set to 3 to reduce the impact of the total amount of training data on model performance.

Table 3 Fine-tuning experiments

| Group | Data size | Components |
|---|---|---|
| A | 2129 | 100% QA |
| B | 2661 | 80% QA + 20% QRA |
| C | 2500 | 60% QA + 40% QRA |
| D | 1364 | 100% QRA |

## 4 Result

### 4.1 BIM-QA dataset

In this study, Revit models of various building types, including shopping malls, office buildings, dormitories, teaching buildings, and museums, were used as the sources for dataset generation. To ensure balanced sample sizes and appropriate component density, all models were partitioned into spatial blocks of predefined sizes. The block size for each model was determined through manual experimentation to maintain a reasonable number of components within each block. Two block sizes, 5 m and 8 m, were adopted in this study, with each block containing approximately 10-15 components. A total of 150 model blocks were obtained.

For each block, defect data were first generated following the method described in **Section 3.2.1**. Subsequently, specific evaluation questions were generated based on the EQS, and reference answers were computed using hard-coded rules, resulting in 22 QA pairs. Meanwhile, both the original and defect-injected BIM data were converted into textual form and output together with the questions and answers, resulting in four corresponding txt files.

Through this process, a total of 3,300 BIM-QA data pairs were generated, providing a unified and structured input for subsequent model evaluation. **Table 7** summarize the block parameters of the different building models and the corresponding quantities of generated data pairs.

Table 7 BIM-QA Dataset Statistics

| Building type | Block size | Number of blocks | Number of BIM-QA data pairs |
|---|---|---|---|
| Shopping mall | 8m | 40 | 880 |
| Office building | 8m | 40 | 880 |
| Dormitory | 5m | 30 | 660 |
| Teaching building | 5m | 20 | 440 |
| Museum | 5m | 30 | 660 |
| Total | | 150 | 3300 |

### 4.2 Evaluation of general LLMs

#### 4.2.1 Overall capability of general LLMs

Building on the evaluation setup in Section 3.3.2, this subsection examines the overall capability of the 11 general LLMs on BIM based design tasks.

(1) Performance of closed-source non-reasoning LLMs

To analyze the overall capabilities of general LLMs, the average scores of the quantitative indicators for all questions were calculated. **Fig. 9** shows the evaluation results of six closed-source non-reasoning LLMs from the ChatGLM and Qwen series, where the three models in each series are sorted by their G-Eval scores in ascending order.

In terms of the format scores, the most capable models—GLM-4-plus and Qwen-plus—achieved 100% compliance with format requirements, while other models had instances of non-compliance. This indicates that the format score can characterize an LLM's ability to follow instructions. For less capable LLMs, prompt learning alone cannot ensure format accuracy.

As for BLEU-1 and BLEU-2, Qwen-max scored higher than Qwen-plus. However, their scores in the ROUGE-L indicator were very close. This suggests that it is difficult to judge an LLM's ability to understand and solve problems based solely on text similarity indicators. Furthermore, all models achieved fairly similar Cos-Sim scores, indicating that there was little difference in semantic directionality in this study, and the Cos-Sim indicator had weak discriminative power. Therefore, in subsequent analyses, differences in Cos-Sim scores will not be considered. And only the results of the ROUGE-L indicator among above mentioned indicators will be presented.

The G-Eval indicator showed significant differences between different models. Manual review revealed that G-Eval scores are the closest to human understanding, as they consider not only semantic similarity but also text similarity, making the evaluation results more valuable for reference. So, G-Eval scores will be focused in subsequent analyses. Interestingly, although developers claim that the Qwen-max model has better overall capabilities than Qwen-plus, in this study and this domain, Qwen-plus achieved a higher G-Eval score.

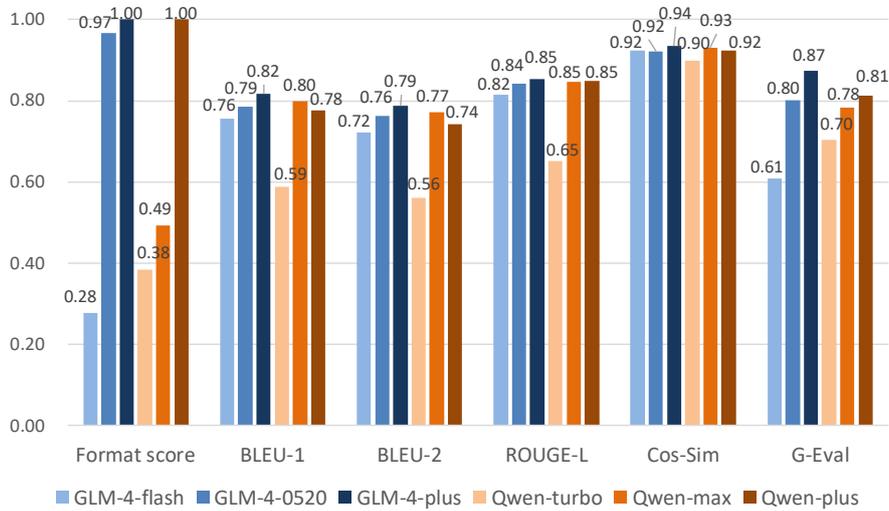

**Fig. 9.** Overall scores of LLMs from ChatGLM and Qwen series

(2) Performance comparison of reasoning and non-reasoning LLMs

To further compare the impact of model reasoning on solving BIM based design tasks,, the study selected the highest-scoring GLM-4-plus and Qwen-plus, along with the DeepSeek-V3 model, to compare with three reasoning models (QwQ-plus, QwQ-32b, and DeepSeek-R1). The results are shown in **Fig. 10**. It can be seen that the overall capabilities of the reasoning models are strong, being inferior to GLM-4-plus but superior to the other two models. Furthermore, although

DeepSeek-V3 and DeepSeek-R1 have the same parameter size, influenced by the reasoning process, the G-Eval score of DeepSeek-R1 significantly improved. In particular, the reasoning model gained a notable increase in the format score. However, the ROUGE-L score of the reasoning model was lower, indicating that while the DeepSeek-R1 model adhered to the format requirements, its output was more variable and quite different from the examples. A manual analysis of specific answers from the DeepSeek-R1 model revealed that a large amount of additional information was added to the answers, leading to low text similarity.

As for the impact of parameter size on reasoning models, in the domain-specific context of this study, the QwQ-32b model with only 32b parameters achieved a higher G-Eval score than the DeepSeek-R1 model (671b). Moreover, both the format score and ROUGE-L score of the DeepSeek-R1 model were significantly lower than those of the other two reasoning models. This suggests that in specific domains, a huger parameter size does not mean better performance, as a cumbersome reasoning process may cause the model to generate redundant information.

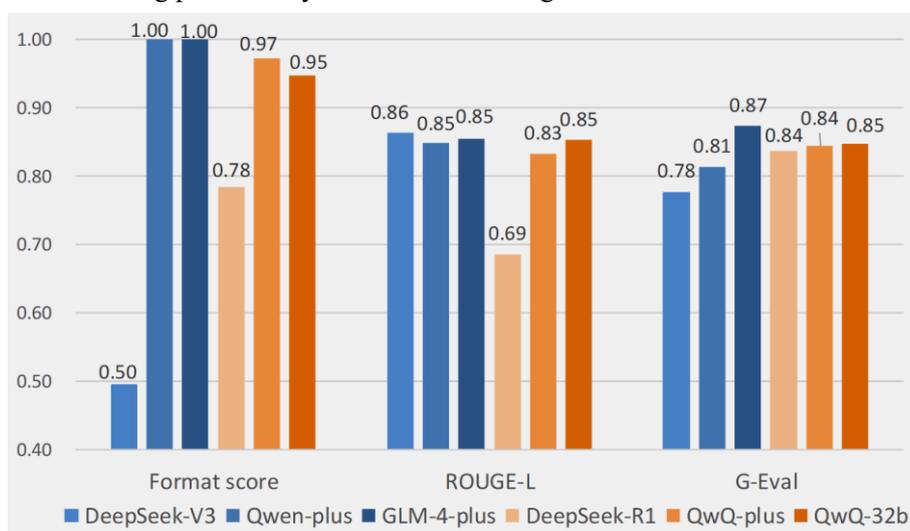

**Fig. 10.** Comparison between reasoning models and non-reasoning models

**4.2.2 Performance on different task types**

To further analyze the capability differences of LLMs in different tasks, this study takes Qwen-plus and the open-source model Qwen2.5-14B-instruct (referred to as the Qwen-14B model for short) as examples to compare their G-Eval scores across different questions. As shown in **Fig. 11,** for the same model, the capabilities in data extraction and statistics are generally stronger than those in calculation and reasoning. This reflects the limitations of the models in mathematical calculation and domain understanding, which aligns with the academic consensus that LLMs are deficient in handling mathematical and logical reasoning problems(Saba, 2024). This limitation is particularly evident in Level 2 reasoning questions, where the model needs to infer floor height from BIM coordinate information without an explicit explanation of this concept, and its performance drops significantly on such items.

For models with different parameter sizes, Qwen-plus and the Qwen-14B model achieve close scores on simple questions, and even the Qwen-14B model with fewer parameters performs better. However, on difficult calculation and reasoning questions, the scores of the Qwen-14B model are generally lower than those of Qwen-plus. This indicates that a larger parameter size, to a certain extent, endows LLMs with the ability to handle complex problems.

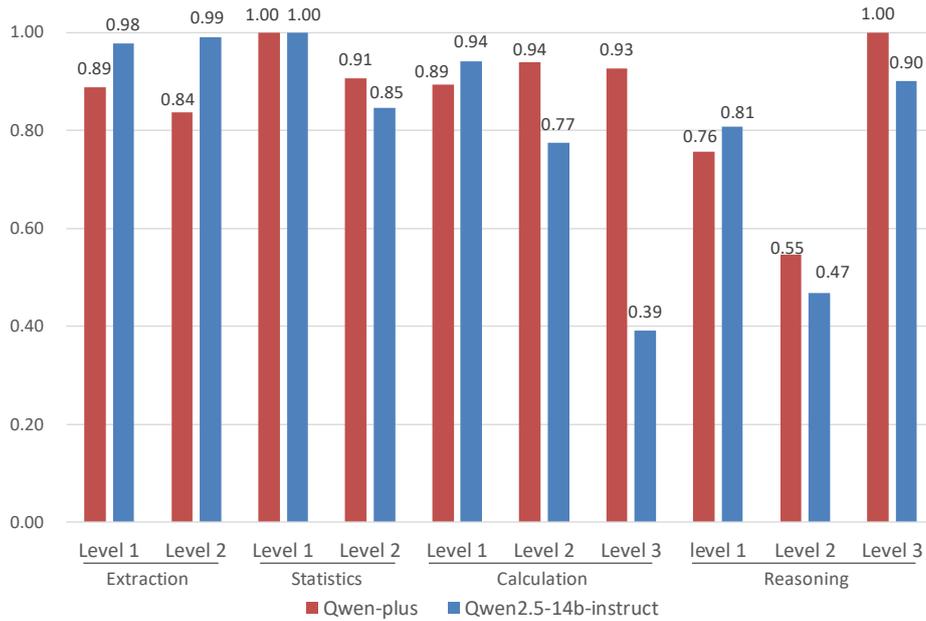

**Fig. 11.** G-Eval scores comparison for general capabilities

As shown in **Fig. 12**, for BIM design review and design detailing tasks, both general LLMs exhibit clear weaknesses, with G-Eval scores generally below 0.8. In almost all cases, the Qwen-14B model performs worse than Qwen-plus, and it even fails to handle some basic compliance checking questions. These results indicate that, without domain-specific fine-tuning, general LLMs still lack sufficient understanding of BIM semantics to reliably solve BIM specific design tasks.

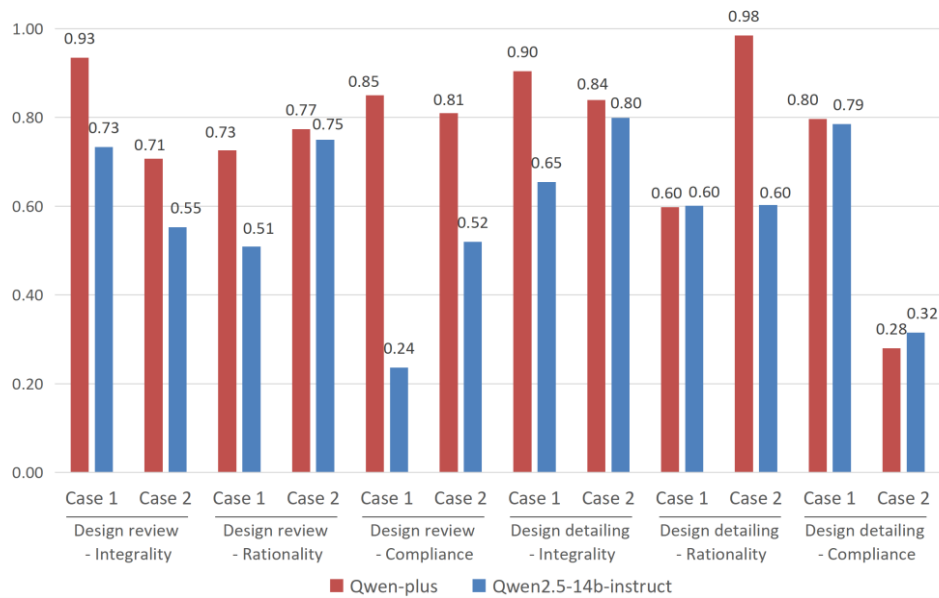

**Fig. 12.** G-Eval scores comparison for BIM Domain-specific understanding capabilities

### 4.2.3 Discussion

The aforementioned experiments demonstrate that the LLM evaluation system incorporating both text similarity and semantic similarity can quantify model performance. Furthermore, the results indicate that reasoning models exhibited stronger overall capabilities than non-reasoning models. However, excessively large parameter sizes can impair performance through redundant reasoning processes. With respect to different tasks, LLMs perform adequately on simple general questions but exhibit deficiencies in complex reasoning problems and domain-specific questions, which collectively underscore the necessity for LLM fine-tuning.

Furthermore, in order to understand the reasons behind the incorrect answers provided by LLMs, below summarizes typical error types based on complete LLM evaluation records, aiming to guide more efficient domain-specific fine-tuning work.

1) Lack of reasoning process

There is a certain randomness in whether LLMs follow the required format when answering. However, answers that include reasoning and analysis processes are often more compliant. As shown in **Fig. 13**, the Qwen-max model often provides final answers directly without reasoning or analysis, which is neither in line with the format requirements nor prone to being incorrect. Moreover, without a reasoning process, it is difficult to analyze the causes of the errors. In contrast, when answering the same question, the Qwen-plus model conducts detailed analysis for component one by one, provides reasons for non-compliance, and gives correct results in accordance with the format requirements. Although for some simple questions, Qwen-max can provide correct results even without a reasoning process, for complex questions, results without reasoning are often wrong. This indicates that guiding LLMs to reason can help them understand questions and requirements, thereby providing correct answers.

> **Question Type:** Compliance identification for construction number
>
> **Qwen-plus: Correct answer          Detailed reasoning process**
> According to the naming rules for component construction numbers, the construction number of a wall must start with "Q", that of a beam must start with "L", and that of a slab must start with "B". We have checked all components with construction numbers in the model one by one.
> **Wall Component Inspection:**
> 1. **id: 342672** - Construction number: Q43 → Compliant.
> 2. **id: 342673** - Construction number: Q43 → Compliant.
> 3. **id: 342676** - Construction number: Q-43 → Non-compliant (contains illegal character "-").
> 4. **id: 342677** - Construction number: Q40 → Compliant.
> ……
> **Summary:**
> The IDs of components with non-compliant construction number naming among the components to be checked are as follows:
> •Wall components: 342676, 342759
> •Beam components: 348188
>
> [Final Answer]:
> The IDs of components with non-compliant construction number among all components to be checked in the model are:
> 342676, 342759, 348188
>
> **Qwen-max: Wrong answer          No reasoning process**
> The IDs of components with non-compliant construction number naming among all components to be checked in the model are as follows:
> 342734, 347861

Fig. 13. Reasoning models are more likely to output in the correct format

2) Limited mathematical calculation ability

As shown in **Fig. 14**, regarding the problem of calculating slab area, although the Qwen-max model understands the question correctly and uses the correct formula, its calculation result is wrong. The 14B model does not provide a calculation formula, and the output area is also incorrect. This shows that even for LLMs with strong comprehensive capabilities, ensuring the accuracy of calculation results is still challenging even when they correctly understand the question.

**Question Type:** Calculation of slab area

**Qwen-max: Correct answer    Detailed calculation but incorrect**
- The ID of the first slab is 349718. It is rectangular, and its area is (23597 - 17597) * (4659 - 1259) / 1000000 = 3.24 square meters.
- The ID of the second slab is 349791. It is rectangular, and its area is (17397 - 15897) * (3059 - 1909) / 1000000 = 0.26 square meters.
- The ID of the third slab is 349800. It is rectangular, and its area is (17397 - 15897) * (4659 - 3259) / 1000000 = 0.32 square meters.
- The ID of the fourth slab is 349811. It is not rectangular.
- The ID of the fifth slab is 349929. It is not rectangular.
- The ID of the sixth slab is 349938. It is rectangular, and its area is (19147 - 17597) * (6159 - 4859) / 1000000 = 1.44 square meters.

The largest rectangular floor slab among these has an ID of 349718 and an area of 3.24 square meters.

**Qwen2.5-14b-instruct: Wrong answer    No calculation process**
- The ID of the first floor slab is 350297. It is rectangular with an area of 2.7 m².
- The ID of the second floor slab is 350306. It is rectangular with an area of 11.88 m².

The largest rectangular floor slab is ID 350306, with an area of 11.88 m².

**Fig. 14.** Mathematical calculation error

3) Misunderstanding of natural semantics

Misunderstandings of the content of questions, examples, and rules can lead to serious errors in LLM answers. **Fig. 15** provides two examples: the Qwen-plus model fails to understand the meaning of parentheses in the example content, resulting in obvious errors in the answer. The GLM-4-flash model does not understand the combination rules of letters and numbers in the naming rules, so although it corrects the error in the letter, it still retains the wrong connection symbol. These issues expose the deficiencies of LLMs in natural semantic understanding. In comparison, more capable models such as GLM-4-plus have more accurate understanding of natural language.

**Question Type:** Calculation of wall length

**Example**
The content after [Final Answer] follows the example below:
The length of the third wall is 900 mm, and the given length data is 900 (1000) mm. The given data is correct (incorrect).

> The parentheses indicate selectable output results, rather than simply copying the original text

**Qwen-plus: Wrong answer**
… …
[Final Answer]:
The length of the first wall is 602 mm, and the given length data is 1100 mm. The given data is correct (incorrect). ✗

**Cannot understand parentheses**

**Question Type:** Compliance repair for construction number

**Naming rules**
Here are naming rules for component construction numbers: A component's construction number consists only of one initial letter followed by numbers. The construction number of a wall must start with "Q", that of a beam must start with "L", and that of a slab must start with "B".

**GLM-4-flash: Wrong answer**
According to the given naming rules for construction numbers, the following modifications should be made to these components:
The construction number of the component with ID 342710 should start with "Q", so it should be modified to: Q-16. ✗
The construction number of the component with ID 347936 should start with "L", so it should be modified to: L37.

**Cannot understand naming rules**

**Fig. 15.** Misunderstanding of natural semantics

4) Lack of domain-specific knowledge

More commonly, LLMs lack the ability to understand professional terms and concepts in the domain-specific fields. As shown in **Fig. 16**, the Qwen-14B model does not understand how to reason about the floor height of a building model and mistakenly believes that the floor height is determined by the floor thickness. Although the reasoning process of LLM suggests that the result goes against common sense of design, it remains stuck in the wrong logic. This is mainly because general models are pre-trained using general corpus and lack domain-specific knowledge. This also illustrates the importance of fine-tuning general LLMs to solve domain-specific problems.

**Question Type:** Reasoning of floor height

**Qwen2.5-14b-instruct: Wrong answer**

Since the upper surfaces of all mentioned slabs are aligned with the elevation of the second floor, and the thickness of each floor slab is 120 mm, it can be inferred that the floor height of the building at least includes the thickness of the slab plus the space height below the slab. Considering that the space height below the floor slab usually includes the height of ceilings or other facilities, but specific values are not provided, we can assume that this space height is close to the thickness of the slab for simplified calculation. Therefore, the floor height of the building is most likely twice the thickness of the slab, that is, 240 mm. However, considering the standard practices in actual architectural design, the floor height is usually much larger than this value, so a more reasonable inference is that the space height below the slab is close to or slightly higher than the thickness of the slab. Based on this, if we set the space height below the slab to be the same as the thickness of the slab, the total floor height is 120 mm + 120 mm = 240 mm, but this is obviously inconsistent with the actual situation; if we set it to be slightly higher than the thickness of the slab, such as 130 mm, the total floor height is 120 mm + 130 mm = 250 mm, which is still on the low side. In view of the common floor height range in architectural practice, a more reasonable assumption is that the space height below the floor slab is approximately between 1.5 and 2 times the thickness of the floor slab, so the floor height of the building is most likely between 120 mm + 180 mm and 120 mm + 240 mm, that is, between 300 mm and 360 mm. To provide a specific answer, we take the median value as the most likely floor height, that is, 330 mm.

**Having common sense but wrong understanding of floor height**

Fig. 16. Incomprehension of domain-specific knowledge

### 4.3 Development of domain-specific LLM

In this part, a domain-specific LLM for BIM-based design, referred to as Qwen-BIM, is developed. Based on the comparative evaluation of general models, Qwen2.5-14B-instruct is chosen as the base model to balance capability and computational cost. On this basis, the model is further trained with carefully curated fine-tuning datasets that combine QA pairs and QRA samples, and different proportions of reasoning data are compared to determine the most effective configuration. The model obtained from the best-performing configuration is named Qwen-BIM and is subsequently used for all domain-specific evaluations.

#### 4.3.1 Selection of base open-source LLM

Fine-tuning is essential for adapting LLMs to BIM-based design tasks, so selecting an appropriate open-source base model is a prerequisite for domain-specific development. Building on the comparative results in Section 4.2, **Fig. 17** compares a series of Qwen2.5 open-source models with different parameter sizes against three closed-source models, namely Qwen-max, Qwen-plus, and Qwen-turbo.

In terms of format scores, the open-source models perform better than Qwen-max and Qwen-turbo, but still fall short of 100% accuracy, which leaves room for improvement through subsequent fine-tuning. From the perspective of ROUGE-L and G-Eval, the 32B and 72B open-source models achieve capabilities comparable to, and in some cases exceeding, Qwen-plus. This suggests that models with moderately smaller parameter sizes can still attain strong performance on BIM-related tasks. However, when the parameter size is too small, overall capability is clearly constrained; the

7B model lags behind the other open-source models on all indicators.

These results indicate that, for BIM design review and detailing tasks, a certain parameter scale is sufficient to guarantee baseline performance, while further gains mainly depend on domain-specific development. Taking into account both capability and the available GPU resources for fine-tuning experiments (96 GB across two NVIDIA RTX A6000 GPUs), this study selects the Qwen2.5-14B-instruct open-source model as the base LLM for subsequent domain-specific training.

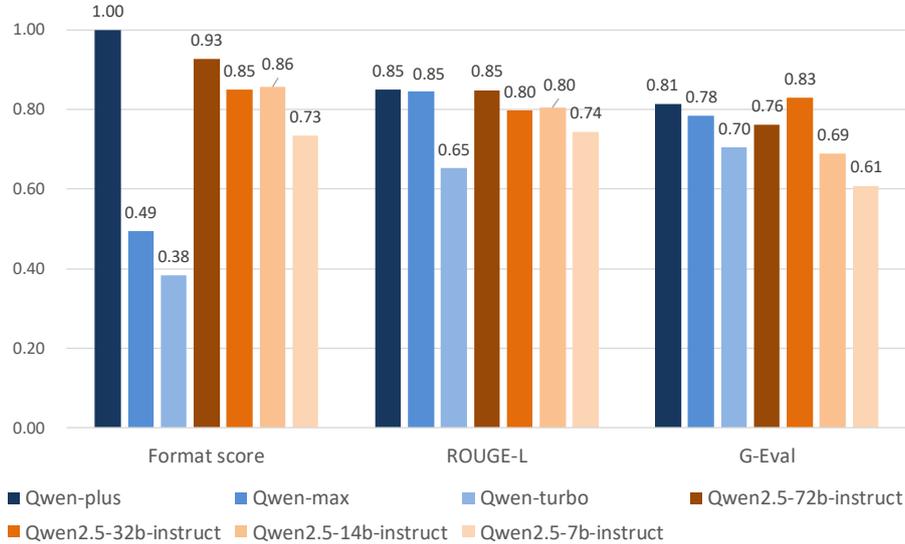

**Fig. 17. Comparison between Qwen series open-source models and closed-source models**

### 4.3.2 Impact of reasoning data on fine-tuning

**Fig. 18** shows the evaluation results of the base LLM (Qwen2.5-14B-instruct) and the four groups of fine-tuned models (Groups A~D in **Table 6**), where the datasets differ in the proportion of QRA reasoning data. In terms of G-Eval performance, all fine-tuned models achieved scores no lower than the base LLM. And, the higher the proportion of reasoning data in the dataset, the better the performance of the fine-tuned model. This proves that guiding LLMs to reason can help them provide more accurate answers. Notably, the fine-tuning effect of the dataset with 100% QRA data was the best, even though its data size was the smallest. This also indicates that in the process of fine-tuning, data quality is more important than data quantity.

As for format scores, there was no obvious pattern in the fine-tuning effects of different datasets. However, Group D (with 100% QRA data) achieved the best performance with an accuracy rate of 90.2%. From the perspective of text similarity, more QRA reasoning data led to lower ROUGE-L scores, which is consistent with the conclusion in Section 4.2.1: reasoning introduces redundant model outputs that are irrelevant to the answer, thereby reducing the text similarity score without affecting the correctness of the answer.

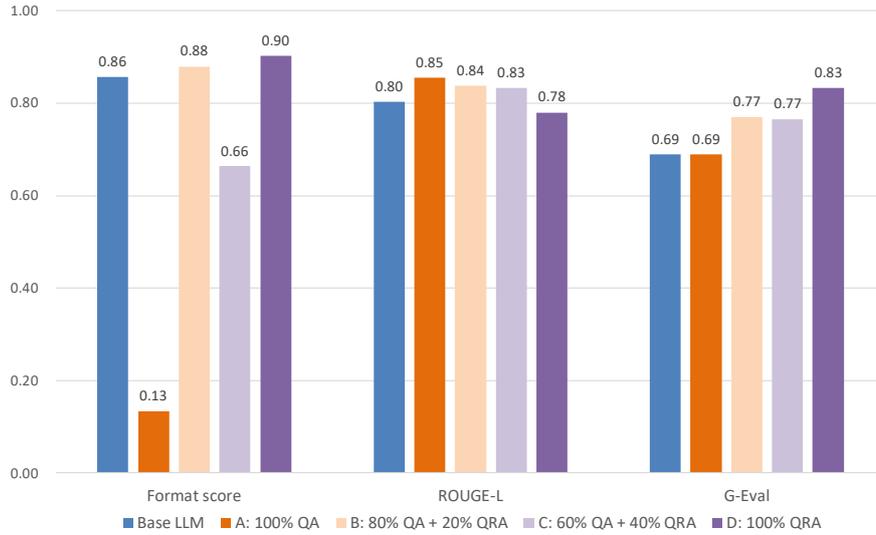

**Fig. 18.** The performance impact of resoning data on model fine-tuning

**4.3.3 Capability of Qwen-BIM**

Compared with the base LLM, as shown in **Fig. 18**, the fine-tuned model of Group D achieved a 5.3% improvement of format score, and a 21.0% improvement of G-Eval score. This model, which showed the best fine-tuning performance, is referred to as the Qwen-BIM LLM in BIM domain. And the following is a more comprehensive evaluation and analysis of Qwen-BIM.

**Fig. 19** compares the performance of Qwen-BIM with that of other general LLMs with different parameter sizes. With 14B parameters, Qwen-BIM outperforms models in the same series with 32B and 72B parameters, and its G-Eval score almost matches that of the DeepSeek-R1 reasoning LLM with 671B parameters. These results indicate that the fine-tuning strategy can substantially enhance the ability of LLMs to solve domain-specific problems related to BIM-based design. They also suggest that, if computational resources permit, applying the same strategy to base models with larger parameter sizes may lead to further performance gains.

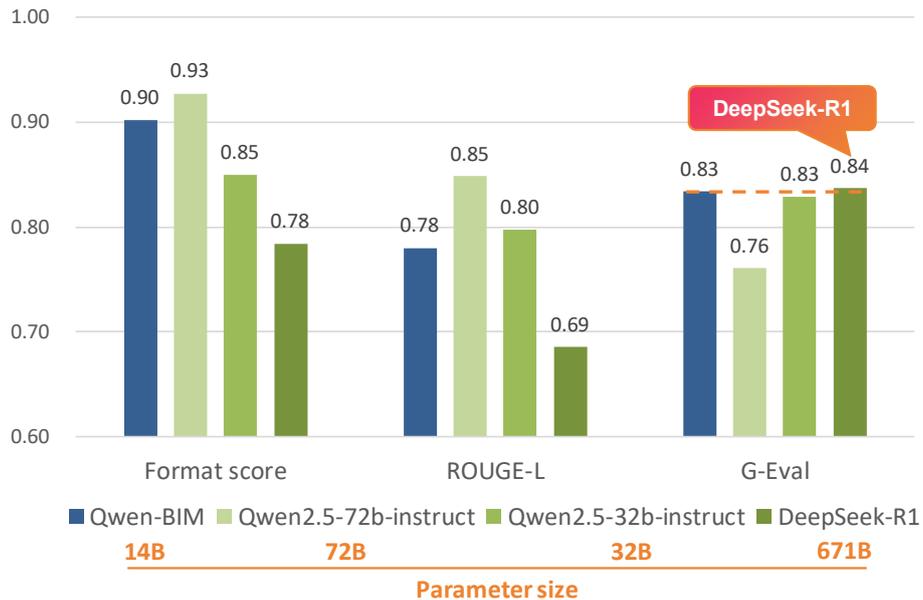

**Fig. 19.** Evaluation of Qwen-BIM and general LLMs with different parameter sizes

The G-Eval scores of the LLM before and after fine-tuning further corroborate the crucial role of fine-tuning. As shown in **Table 8**, fine-tuning increases the overall G-Eval score from 0.689 to 0.834, with a moderate improvement on general tasks (from 0.810 to 0.874) and a notably larger improvement on domain-specific tasks (from 0.588 to 0.801). **Fig. 20(a)** plots the G-Eval scores of Qwen-BIM and the base LLM (Qwen2.5-14B-instruct) across all question types, while **Fig. 20(b)** presents the score differences obtained by subtracting the base LLM scores from those of Qwen-BIM, where green bars indicate performance gains and red bars indicate declines.

(1) Overall performance distribution

As illustrated in **Fig. 20(b)**, Qwen-BIM achieves higher G-Eval scores than the base LLM on 13 question types. Most of these gains exceed 0.1, and the average improvement across these types reaches 0.28. By contrast, Qwen-BIM performs slightly worse on 8 question types, where most deviations lie between 0 and -0.1. The average decrease in these cases is only 0.06, indicating that even where performance drops, Qwen-BIM remains broadly comparable to the base LLM. Overall, the net effect of fine-tuning is a substantial and stable enhancement of performance.

(2) Performance on general tasks

The question types showing decreased scores after fine-tuning are mainly concentrated in extraction, statistics, and calculation tasks for general questions. These tasks are already handled relatively well by the base LLM, so the room for further improvement is limited, and small fluctuations introduced by fine-tuning may slightly affect text patterns and thus G-Eval scores. Nevertheless, the magnitude of decline on these tasks is modest, and the basic capabilities of Qwen-BIM on general questions remain close to those of the base model.

(3) Performance on domain-specific tasks

For complex reasoning tasks and BIM domain-specific tasks, Qwen-BIM exhibits clear performance superiority. The results show that, for domain-specific questions, Qwen-BIM outperforms the base LLM on 75% of question types, and the average G-Eval improvement over the base model reaches 0.21. The gains are particularly evident on design review and design detailing tasks that require the model to interpret BIM semantics, apply design common sense, and reason under specification constraints.

In summary, fine-tuning markedly enhances the comprehensive capability of the general LLM and is critical for solving BIM-based domain-specific problems. The fine-tuning dataset constructed in this study enables the model to better handle design review and design detailing tasks in BIM-based design, thereby demonstrating the effectiveness of the proposed domain-specific LLM development approach.

Table 8 G-Eval comparison between Qwen-BIM and base model

|  | Overall | General tasks | Domain-specific tasks |
| --- | --- | --- | --- |
| Base LLM | 0.689 | 0.810 | 0.588 |
| **Qwen-BIM** | **0.834** | **0.874** | **0.801** |
| Improvement | 0.145 | 0.064 | 0.212 |

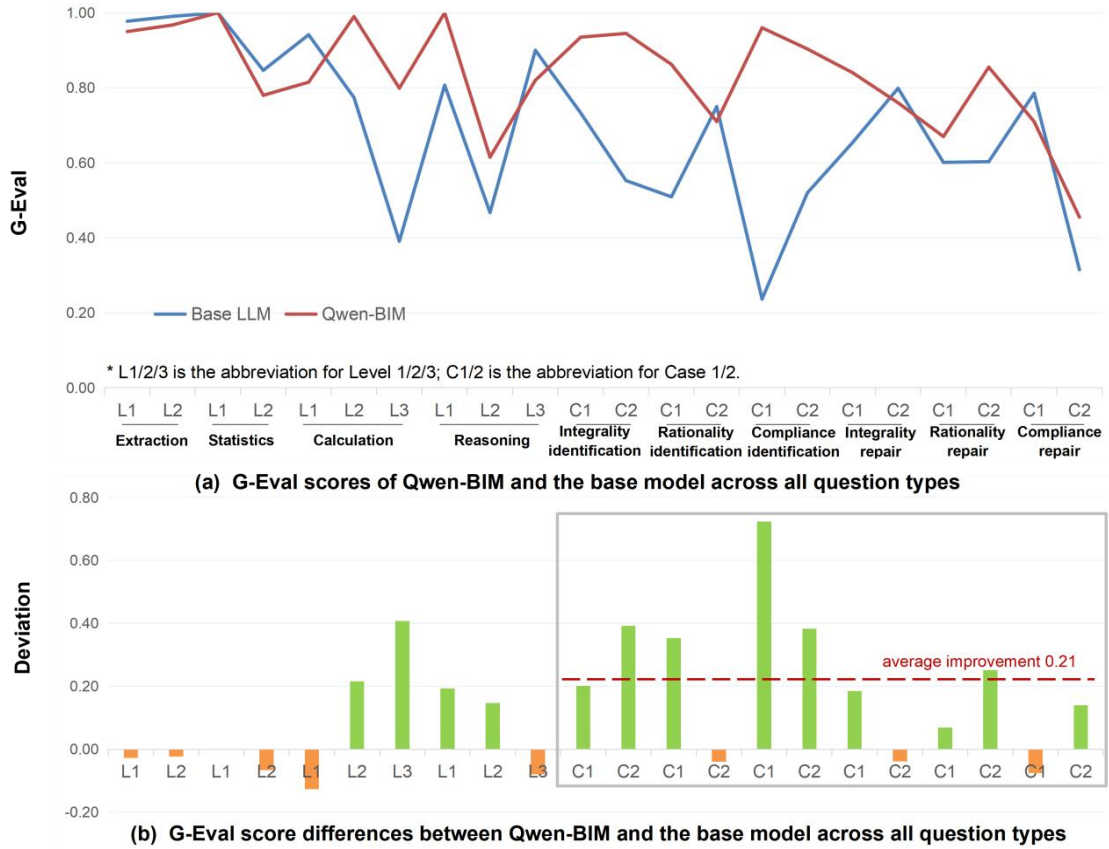

**Fig. 20.** G-Eval comparison between Qwen-BIM and base model across all question types

## 5 Conclusion

While research and applications of LLM technology are growing rapidly across industries, the civil engineering field features high professional barriers, resulting in relatively slow progress in LLM-based applications. This study attempts to integrate LLM technology into BIM-based design field, thus proposing a comprehensive domain-specific LLM evaluation and fine-tuning method. Firstly, a method for generating textual data from BIM is proposed. On this basis, the capability requirements of LLMs are analyzed, and a series of question-answer sets are produced from BIM to form the evaluation datasets. Then quantitative indicators are proposed to together establish a comprehensive LLM evaluation benchmark. Finally, a series of LLM evaluation experiments have been implemented. And according to the evaluation results, a domain-specific dataset is constructed to realize the fine-tuning of general LLM. The evaluation and fine-tuning results show that 1) The proposed domain-specific benchmark can effectively and comprehensively assess the capabilities of LLMs; 2) General LLMs possess general capabilities to solve simple problems but are incompetent in handling domain-specific issues; 3) Fine-tuning based on the constructed domain-specific dataset can effectively improve model performance, especially in terms of domain-specific capabilities; 4) The more high-quality reasoning data there is, the better the performance of the fine-tuned model. The optimal Qwen-BIM fine-tuned model achieves an average increase of 21.0% in G-Eval score compared to the base LLM model, and with 14B parameters, it can achieve performance comparable to that of general LLMs with 671B parameters.

For the first time, this study proposes a comprehensive domain-specific benchmark and a high-quality domain-specific dataset, developing LLM for the field of BIM-based design. The constructed LLM evaluation and fine-tuning methods, including the domain-specific benchmark and dataset, have transfer value and can provide references for developing BIM-related LLMs in various fields.

Future research should focus on expanding the benchmark to cover richer BIM-based design, construction, and operation tasks, deploying the model in more practical application scenarios, exploring more powerful base LLMs and fine-tuning strategies, and conducting large-scale validation in real engineering projects.

**CRediT authorship contribution statement**

**Jia-Rui Lin:** Conceptualization, Supervision, Methodology, Writing - review & editing, Funding acquisition, Project administration. **Yun-Hong Cai:** Writing - original draft, Writing - review & editing, Methodology, Data curation, Formal analysis, Visualization. **Xiang-Rui Ni:** Writing - original draft, Methodology, Validation, Investigation, Data curation. **Shaojie Zhou:** Writing - original draft, Writing - review & editing, Methodology, Visualization, Investigation. **Peng Pan:** Conceptualization, Supervision, Project administration, Funding acquisition, Writing - review & editing.

**Data availability**

Data will be made available on request.

**Acknowledgments**

This research was conducted with the supports of the National Key R&D Program of China (Project No. 2023YFC3804600) and the National Natural Science Foundation of China (Project No. 52378306).